# Evaluating BERT and ParsBERT for Analyzing Persian Advertisement Data

Ali Mehrban[1] and Pegah Ahadian[2]

[1]Newcastle University, Newcastle upon Tyne, UK
[2]Dept of Computer Science, Kent State University, Kent, USA

## Abstract

*This paper discusses the impact of the Internet on modern trading and the importance of data generated from these transactions for organizations to improve their marketing efforts. The paper uses the example of Divar, an online marketplace for buying and selling products and services in Iran, and presents a competition to predict the percentage of a car sales ad that would be published on the Divar website. Since the dataset provides a rich source of Persian text data, the authors use the Hazm library, a Python library designed for processing Persian text, and two state-of-the-art language models, mBERT and ParsBERT, to analyze it. The paper's primary objective is to compare the performance of mBERT and ParsBERT on the Divar dataset. The authors provide some background on data mining, Persian language, and the two language models, examine the dataset's composition and statistical features, and provide details on their fine-tuning and training configurations for both approaches. They present the results of their analysis and highlight the strengths and weaknesses of the two language models when applied to Persian text data. The paper offers valuable insights into the challenges and opportunities of working with low-resource languages such as Persian and the potential of advanced language models like BERT for analyzing such data. The paper also explains the data mining process, including steps such as data cleaning and normalization techniques. Finally, the paper discusses the types of machine learning problems, such as supervised, unsupervised, and reinforcement learning, and the pattern evaluation techniques, such as confusion matrix. Overall, the paper provides an informative overview of the use of language models and data mining techniques for analyzing text data in low-resource languages, using the example of the Divar dataset.*

## 1. Introduction

The exchange of goods and services between two individuals or organizations for payment is known as trade. Nowadays, people differentiate between two types of trading: traditional trading prior to the advent of the Internet and modern trading after it. The way we trade has massively changed over time. The advancement of the Web and the utilization of different software programs and mobile applications have fundamentally impacted individuals' lifestyles by empowering Internet shopping and smoothing out the method involved with placing orders for services and products from the solace of their homes.

A person can browse an app or website and find anything they are looking for with only a few clicks on a laptop or a few taps on a smartphone when connected to the Internet. The item is delivered to their door within a few hours after they research pricing, decide to buy, and pay with a credit card. Similar to buying, selling has grown quicker and more convenient as consumers can now post ads directly on a variety of websites and connect with clients directly without the use of middlemen. These transactions produce a lot of data, which needs to be understood because it provides important information. This information can help organizations improve their marketing efforts in order to boost sales by giving them a valuable understanding of consumer attitudes, preferences, and trends.

Briefly, with the rise of the Internet, the variety of software and applications has transformed the way we do trade, making it faster, more convenient, and accessible to people worldwide. The data generated from these transactions can provide valuable understanding and superiority, allowing companies to optimize their strategies and increase their revenue.Divar is an online marketplace established for buying and selling new and used products and services in Iran. It was created to provide a platform for transactions without intermediaries that allows people to buy and sell directly. The challenge for participants in the competition was to predict what percentage of a car sales ad would be published on the Divar website. The dataset for the competition provides a rich source of Persian text data. Persian is a language spoken in countries such as Iran and Afghanistan and is written from right to left. The Persian alphabet is a derivation of the Arabic script and is used officially in Iran and Afghanistan, while in Tajikistan it is written in the Tajik alphabet,which is derived from the Cyrillic script. There is a vast amount of Persian written material available, including books, newspapers, scientific papers, and online pages.

To process and analyze the Persian text data from the Divar dataset, the authors of the paper used the Hazm library, which is a Python library specifically designed for digesting Persian text. They also utilized two state-of-the-art language models: mBERT and ParsBERT. BERT is a transformer-based model that was originally developed for the English language but is also applicable to other languages. It is known as a multilingual model because it can be trained in languages other than English and then fine-tuned for specific tasks. ParsBERT is a monolingual model developed specifically for the Persian language.

The paper's primary objective is to compare the performance of mBERT and ParsBERT on the Divar dataset. The authors present their findings in several sections. They begin by providing some background on data mining and the two language models. They then delve into the Divar dataset and examine its composition and statistical features. The authors also provide details on their fine-tuning and training configurations for both approaches. Finally, they present the results of their analysis, which sheds light on the strengths and weaknesses of the two language models when applied to Persian text data. Overall, the paper offers valuable insights into the challenges and opportunities of working with low-resource languages such as Persian and the potential of advanced language models like BERT for analyzing such data.

## 2. PRELIMINARIES

Data mining is a cutting-edge technology that aids businesses in extracting efficient insights from their huge amounts of data. It distinguishes crucial patterns and structures that may not be immediately recognizable from straightforward queries or reports by utilizing advanced algorithms and statistical analysis methods. This helps businesses find information that is hidden in their data. This information can be used to make important business decisions like making new products, marketing plans, and managing customer relationships.

### 2.1. Data mining process steps

- **Data Cleaning**
  Replace missing values using methods likes mean, median, most frequent, and constant values.

- **Normalization**
  To minimize data duplication and avoid issues such as insertion, update, and deletion anomalies, we utilize normalization techniques to bring the data into a specified range. Scaling is applied to keep it within the desired range if we take it as a variable.

- **Min-Max Normalization**

$$v'_i = \frac{v_i - min_A}{max_A - min_A}(new\_max_A - new\_min_A) + new\_min_A$$

$$v'_i = \frac{v_i - \bar{A}}{\sigma_A}$$

- z-score normalization:
- normalization by decimal scaling

## 2.2. Type of Problems

Supervised, Unsupervised, and Reinforcement learning are the three main subtypes of machine learning. In supervised learning, a dataset with labels is available, and the goal is to train a model to anticipate the label of new data that is not known. A binary classification problem is one where the objective is to divide data into two classes. Our data is labeled, and the aim is to divide it into two classes, so it fits into the supervised learning and binary classification categories [1].

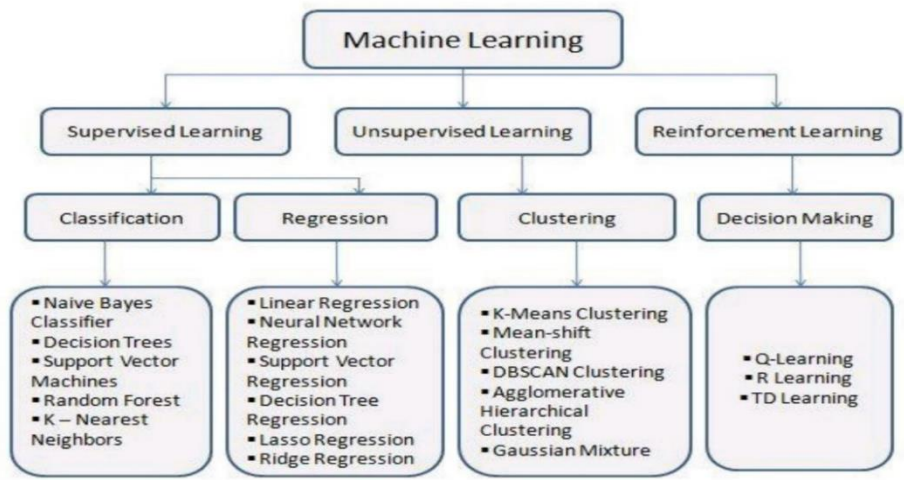

Fig. 1. Subfields and subsections of machine learning

## 2.3. Pattern Evaluation

**Confusion Matrix**

| Predict class / True class | Positive | Negative |
|---|---|---|
| Positive | TP | FN |
| Negative | FP | TN |

Fig. 2. Confusion matrix, shown with totals for positive and negative tuples.

True positives (TP) were cases when the classifier classed the data correctly as positive, whereas true negatives (TN) are scenarios where the classifier classified the data correctly as negative. False positives (FP) and false negatives (FN) are terminologies used to describe circumstances where the results were incorrectly classified as either positive or negative. Two measures, recall, and precision, are utilized to evaluate the forecast coverage and accuracy of the classification system. The F measure, also called the $F_1$ score or F-score, is a way to combine precision and recall into a single measure. They are defined as:

| Measure | Formula |
|---|---|
| accuracy, recognition rate | $\frac{TP+TN}{P+N}$ |
| error rate, misclassification rate | $\frac{FP+FN}{P+N}$ |
| sensitivity, true positive rate, recall | $\frac{TP}{P}$ |
| specificity, true negative rate | $\frac{TN}{N}$ |
| precision | $\frac{TP}{TP+FP}$ |
| $F$, $F_1$, F-score, harmonic mean of precision and recall | $\frac{2 \times precision \times recall}{precision + recall}$ |
| $F_{\beta}$, where $\beta$ is a non-negative real number | $\frac{(1+\beta^2) \times precision \times recall}{\beta^2 \times precision + recall}$ |

Fig. 3. Evaluation measures

## 2.4. Mining Text Data

Text mining is a field that involves various areas such as information retrieval, data mining, machine learning, statistics, and computational linguistics. A considerable amount of information is saved in text format, including news articles, technical papers, books, digital libraries, email messages, blogs, and web pages. As a result, text mining has become a highly active research area. The primary objective of text mining is to extract valuable information from text. It is the process of converting unstructured text data into structured machine-readable data to discover hidden patterns or knowledge discovery databases from the text (KDT). Text mining involves machine learning-supported analysis of textual data [2], [3], [4].

## 2.5. History of BERT and ParsBERT

Jacob Devlin and his colleagues from Google developed BERT and released it in 2018. Google announced in 2019 that it started using BERT in its search engine and by the end of 2020, BERT was being utilized in nearly all English-language queries[6].

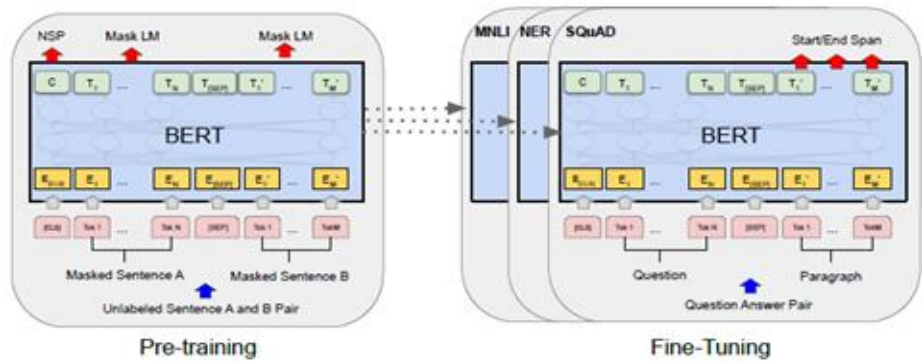

Fig. 4. Overall pre-training and fine-tuning procedures for BERT

The ParsBERT model, which was presented in [5], is a language model for a single language that employs Google's BERT architecture. It has been pre-trained on a large Persian corpus comprising over 3.9 million documents, 73 million sentences, and 1.3 billion words, encompassing various writing styles from several subjects such as science, novels, and news[7],[8].

## 3. INITIAL EXPLORATION OF DIVAR DATA

The dataset is available in the parquet format and its training size is 214Mb. It contains four columns named "post_id", "post_data", "review_label", and "reject_reason_id", and has a total of 540362 rows.

| | post_id | post_data | review_label | reject_reason_id |
|---|---|---|---|---|
| 0 | cb000599-2ee2-42c1-9f0e-32cfeb940398 | {'body_status': 'witout-color', 'brand': '\u06... | accept | 0 |
| 1 | 12063741-6634-444b-befa-0be4c95c2b42 | {'body_status': 'witout-color', 'brand': '\u06... | reject | 13 |
| 2 | 81c93119-5c06-412f-80aa-363ddb0ebc33 | {'body_status': 'witout-color', 'brand': '\u06... | accept | 0 |
| 3 | b5a5bfa7-03be-408b-b4d9-bca26c0ca59b | {'brand': '\u0637\u0627\u06cc\u0631', 'brand_m... | accept | 0 |
| 4 | 5414e920-0faf-44a8-9853-0b03d66e9e2a | {'body_status': 'intact', 'brand': '\u067e\u06... | reject | 12 |
| ... | ... | ... | ... | ... |
| 540357 | 57d8585a-762d-49df-a1ea-fde6ae63e35b | {'brand': '\u0637\u0627\u06cc\u0631', 'brand_m... | accept | 0 |
| 540358 | 7bace186-9fc3-450b-9c23-3a109fa1f455 | {'body_status': 'some-scratches', 'brand': '\u... | reject | 145 |
| 540359 | d8014824-d3e7-4a0a-9863-df11021f23d4 | {'body_status': 'some-scratches', 'brand': '\u... | accept | 0 |
| 540360 | ee2bdfaf-773e-430e-9e04-cc250e7a27c6 | {'body_status': 'witout-color', 'brand': '\u06... | accept | 0 |
| 540361 | 8b88abae-ad36-4824-b19d-161ffc23ce77 | {'body_status': 'two-spots-paint', 'brand': '\... | accept | 0 |

540362 rows × 4 columns

Fig. 5. df=pq.read_table(source=" DMC-Train.parquet").to_pandas()

The final column provides additional details on why ads were either accepted or rejected. The distribution of the labels is not equal, indicating that there are more instances of one label than the others[9].

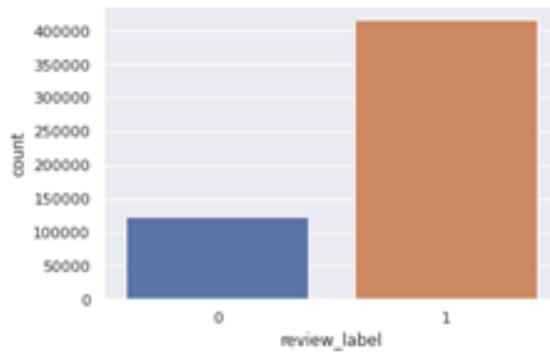

Fig. 6. review_label column

The column 'post_data' contains values in dictionary format which are converted into a dictionary using Json. The columns added by the dictionary are merged with the previous column. The following provides a brief overview of each column in the dataset:

```
dffff.columns

Index(['post_id', 'post_data', 'review_label', 'reject_reason_id',
       'body_status', 'brand', 'brand_model', 'category', 'color',
       'description', 'document', 'gearbox', 'new_price', 'post_type',
       'selling_type', 'third_party_insurance_deadline', 'title', 'usage',
       'year', 'options'],
      dtype='object')
```

Fig. 7. Column's titles

The training dataset comprises 20 columns that contain both textual and numerical values.

| | post_id | post_data | review_label | reject_reason_id | body_status | brand | brand_model | category | color |
|---|---|---|---|---|---|---|---|---|---|
| 0 | rb000599-2ee2-42c1-9f0e-32cfeb840288 | {'body_status': 'witout-color', 'brand': '\u06... | 1 | 0 | witout-color | Tiba تیبا | Tiba Sedan SX | light | سفید |
| 1 | 12063741-6634-444b-befa-0be4c95c2b42 | {'body_status': 'witout-color', 'brand': '\u06... | 0 | 13 | witout-color | Lifan لیفان | Lifan X60 manual | light | 9... |
| 2 | 81c92119-5c06-412f-80aa-363ddcb0ebc33 | {'body_status': 'witout-color', 'brand': '\u06... | 1 | 0 | witout-color | Peugeot پژو F=0 | Peugeot 405 SLX | light | خاکستری |
| 3 | b5a56fa7-03be-408b-b4d9-bca26c0ca59b | {'brand': '\u0633\u0627\u06cc\u0631', 'brand_m... | 1 | 0 | NaN | سایر | Dena basic 1700cc | light | سفید |
| 4 | 3414e920-dfaf-44a8-9d52-0b03d66e9e2a | {'body_status': 'intact', 'brand': '\u067e\u06... | 0 | 12 | intact | Peugeot پژو پژو | Peugeot 206 SD V8 | light | م... |

Fig. 8. df. head ()

Fig. 9. df. head()

Fig. 10. df. info ()

| | review_label | reject_reason_id | new_price | third_party_insurance_deadline | usage |
|---|---|---|---|---|---|
| count | 540362.000000 | 540362.000000 | 4.022780e+05 | 420933.000000 | 540344.000000 |
| mean | 0.770539 | 21.240174 | 2.076257e+08 | 7.889914 | 105651.165735 |
| std | 0.420487 | 50.635343 | 7.809686e+08 | 3.465773 | 115980.108330 |
| min | 0.000000 | 0.000000 | 1.000000e+06 | 1.000000 | 0.000000 |
| 25% | 1.000000 | 0.000000 | 6.300000e+07 | 5.000000 | 290.000000 |
| 50% | 1.000000 | 0.000000 | 1.120000e+08 | 8.000000 | 70000.000000 |
| 75% | 1.000000 | 0.000000 | 1.930000e+08 | 11.000000 | 180000.000000 |
| max | 1.000000 | 163.000000 | 5.000000e+10 | 12.000000 | 500000.000000 |

Fig. 11. df. describe ()

Fig. 12. df. describe (include=['0'])

Fig. 13. df. describe ()

Fig. 14. df. isnull().sum()

For the purposes of this paper, we will only focus on the text and review_label columns. We randomly split the dataset, with a focus on the class with fewer labels, the 0 class.

Examining the Distribution of Labels in Comments - Review_Label:

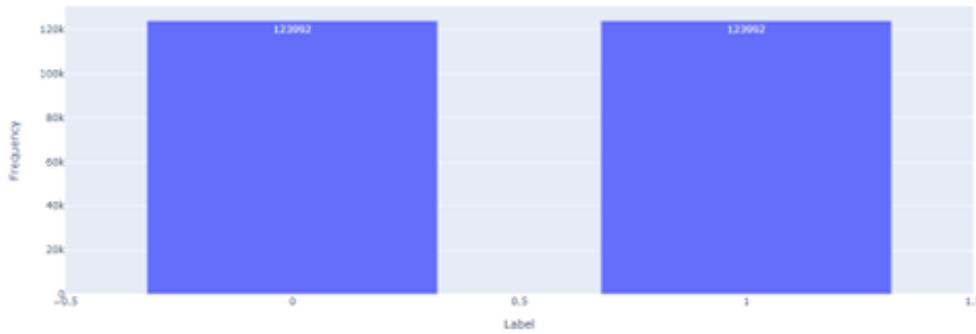

Fig.15. SNS Plotting Divar Data. Distribution of label within comments

## 4. CONFIGURATIONS

This paper focuses solely on the "text" and "review_label" columns.

### 4.1. Train, Validation, Test Split:

The dataset was split into training, validation, and test sets with a ratio of 0.1 for both validation and test sets. Sklearn's "train_test_split" function was used for splitting the dataset while ensuring that the label distribution was preserved.

### 4.2. Hyperparameters and Model:

TensorFlow was used for the same processes, with "Bert-base-multilingual-cased" being used for tokenization and pre-training. The hyperparameters used were as follows:

max_len=128, train_batch_size=16,valid_batch_size=16, epochs=3, Every_epoch=1000, learning_rate=2e5, clip=0.0, optimizer=adam, metric=SparseCategoricalAccuracy('accuracy'), loss=SparseCategoricalCrossentropy, and model= TFBertForSequenceClassification.

"HooshvareLab/bert-fa-base-uncased" was used for tokenization and pre-training, with the same hyperparameters and loss function being used for every task.

### 4.2.1.Fine-Tuning Setup:

For fine-tuning the two models discussed in section II on the Divar dataset explained in section III, we utilized the Adam optimizer with a batch size of 4 and trained for 3 epochs, including 1000 warm-up steps. For Seq2Seq ParsBERT, we set the learning rate to $5e - 5$ [10].

**4.2.2.Training the model by fitting:**

```
Epoch 1/3
12554/12554 [==============================] - 3929s 311ms/step - loss: 0.4399 - accuracy: 0.7913 - val_loss: 0.40
91 - val_accuracy: 0.8106
Epoch 2/3
12554/12554 [==============================] - 3899s 311ms/step - loss: 0.3925 - accuracy: 0.8219 - val_loss: 0.38
87 - val_accuracy: 0.8254
Epoch 3/3
12554/12554 [==============================] - 3889s 310ms/step - loss: 0.3714 - accuracy: 0.8350 - val_loss: 0.38
78 - val_accuracy: 0.8277
FINAL ACCURACY MEAN:  0.8211970660027832
CPU times: user 2h 32min 19s, sys: 21min 30s, total: 2h 53min 49s
Wall time: 3h 15min 17s
```

Fig. 16. Model Fitting (ParsBERT)

```
Epoch 1/3
2022-09-07 15:45:33.508017: I tensorflow/compiler/mlir/mlir_graph_optimization_pass.cc:185] None of the MLIR Opti
mization Passes are enabled (registered 2)
12554/12554 [==============================] - 3574s 283ms/step - loss: 0.4364 - accuracy: 0.7929 - val_loss: 0.3
979 - val_accuracy: 0.8180
Epoch 2/3
12554/12554 [==============================] - 3550s 283ms/step - loss: 0.3787 - accuracy: 0.8310 - val_loss: 0.4
200 - val_accuracy: 0.8276
Epoch 3/3
12554/12554 [==============================] - 3569s 284ms/step - loss: 0.3352 - accuracy: 0.8561 - val_loss: 0.4
342 - val_accuracy: 0.8235
FINAL ACCURACY MEAN:  0.8230212728182474
CPU times: user 2h 14min 42s, sys: 13min 29s, total: 2h 28min 11s
Wall time: 2h 58min 12s
```

Fig. 17.  Models Fitting (mBERT)

# 5. CONCLUSION

To assess the effectiveness of the two structures proposed in this article, we fine-tune both models on the Divar dataset and evaluate them on a binary classification downstream task.

```
1550/1550 [==============================] - 175s 113ms/step - loss: 0.3846 - accuracy: 0.8299

Evaluation: [0.38459542393684387, 0.829872190952301]

              precision    recall  f1-score   support

           0       0.87      0.78      0.82     12399
           1       0.80      0.88      0.84     12400

    accuracy                           0.83     24799
   macro avg       0.83      0.83      0.83     24799
weighted avg       0.83      0.83      0.83     24799

F1: 0.8294321994155076
```

Fig. 18. Evaluation (ParsBERT)

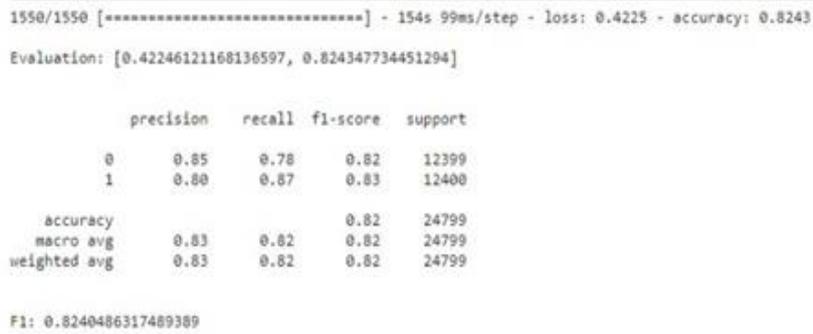

Fig. 19. Evaluation (mBERT)

The table below displays the results achieved on the Divar dataset, indicating that ParsBERT and BERT demonstrate comparable performance in terms of accuracy and F1 score.

This table compares the performance of ParsBert model to that of the Multilingual BERT model:

Table 1. ParsBERT vs mBERT

| Model | Divar Data | |
|---|---|---|
| | *Accuracy* | *F1* |
| ParsBERT | 0.8243 | 0.8240 |
| MultilingualBERT | 0.8299 | 0.8294 |

This study evaluated the processing of Persian text data from the Divar dataset using two cutting-edge language models, mBERT and ParsBERT, and compared their processing abilities. To process and analyze the data, the authors made use of the Python package Hazm, which was created especially for consuming Persian text. They looked into the statistical characteristics of the dataset and presented their training and fine-tuning settings for both methods. Their investigation revealed that, in terms of accuracy, F1-score, and recall, ParsBERT has performed better than mBERT. The results of this study show the potential of cutting-edge language models, like BERT, in the study of low-resource languages like Persian.

This paper has offered helpful details regarding the difficulties and possibilities of dealing with low-resource languages like Persian in the context of data mining. The authors demonstrated how language models, such as BERT, can enhance the precision and effectiveness of such languages' natural language processing tasks. The study's findings show that when applied to Persian text data, ParsBERT, a monolingual model created exclusively for the Persian language, performed better than mBERT, which is a multilingual model.

In conclusion, this study has demonstrated the enormous potential of cutting-edge language models like ParsBERT for processing and studying Persian text data. Businesses and organizations who work with Persian language data should take note of this because it can assist them gain analytical information and improve their tactics. Additionally, this study highlights the need of creating language-specific models for languages with limited resources because they can greatly enhance the precision and effectiveness of activities involving natural language processing. Overall, this study adds to the expanding body of knowledge about low-resource language natural language processing and offers a useful framework for further research in this field.

# AUTHORS

**Ali Mehrban** IEEE Professional member MSc in Communications and Signal Processing from Newcastle University, UK Senior Network Engineer and Researcher at Aryafan Co. Focused on ML and DL applications in network optimization, performance monitoring and resource management

**Pegah Ahadian** PhD. Student in Computer Science at Kent State University, Ohio, USA ACM member Graduate Research Assistant Focused on NLP, ML and DL applications